\documentclass[runningheads,a4paper]{llncs}

\usepackage{amssymb}
\usepackage{graphicx}

\usepackage{url}

\usepackage[normalem]{ulem}
\useunder{\uline}{\ul}{}
\usepackage{multirow}
\usepackage[spanish,es-tabla,activeacute]{babel}

\urldef{\mailsa}\path|{hjthompson, mesapino, ferretti}@unsl.edu.ar|
\urldef{\mailb}\path|merreca@email.unsl.edu.ar|
\newcommand{\keywords}[1]{\par\addvspace\baselineskip
\noindent\keywordname\enspace\ignorespaces#1}
\renewcommand{\keywordname}{{\bfseries Palabras claves:}}

\begin{document}

\mainmatter  

\title{Hacia la interpretabilidad de la detección anticipada de riesgos de depresión utilizando grandes modelos de lenguaje\thanks{Presentado en 30° Congreso Argentino de Ciencias de la Computación (CACIC 2024), La Plata, Argentina. Libro de Actas, pp. 72–81.}}

\titlerunning{Hacia la interpretabilidad de la detección anticipada de riesgos de depresión utilizando LLMs}


\author{Horacio Thompson\inst{1,2}%
\and Maximiliano Sapino \inst{1,2}
\and Edgardo Ferretti \inst{1}
\and\\Marcelo Errecalde\inst{1}
}

\institute{Universidad Nacional de San Luis, San Luis, Argentina \and
Consejo Nacional de Investigaciones Científicas y Técnicas (CONICET) \\
\mailsa\\\mailb}

%


\maketitle
\vspace{-0.4cm}

\begin{abstract}
La Detección Anticipada de Riesgos (DAR) en la Web consiste en identificar usuarios en riesgo tan pronto como sea posible. Aunque los Grandes Modelos de Lenguaje (LLMs) han demostrado resolver eficientemente diversas tareas lingüísticas, es importante evaluar su capacidad de razonamiento en dominios específicos. En este trabajo, proponemos un método para resolver la DAR de depresión utilizando LLMs sobre textos en español, cuyas respuestas puedan ser interpretadas por humanos. Definimos un criterio de razonamiento para analizar usuarios a través de un especialista, aplicamos \emph{in-context learning} al modelo \emph{Gemini}, y evaluamos su desempeño de manera cuantitativa y cualitativa. Los resultados muestran que es posible obtener predicciones precisas sustentadas por razonamientos explicativos, permitiendo una comprensión más profunda de la solución. 
Nuestro enfoque 
 ofrece nuevas perspectivas para abordar los problemas de DAR aprovechando el poder de los LLMs.
\keywords{
Inteligencia artificial, grandes modelos de lenguaje, interpretabilidad, detección anticipada de riesgos, salud mental
}
\end{abstract}

\vspace{-6mm}

\section{Introducción}
La Detección Anticipada de Riesgos (DAR) en la Web consiste en reconocer de forma correcta usuarios que est\'an en riesgo tan pronto como sea posible. Para cada usuario de una red social, se dispone de una lista de publicaciones (posts) y el objetivo es emitir una alarma de riesgo utilizando el menor número posible de posts. Esto incorpora una complejidad adicional a los problemas de clasificación estándar, ya que no sólo es necesario garantizar la precisión, sino también asegurar la rapidez en la toma de decisiones. Si bien la precisión es esencial en la DAR, la importancia de la rapidez incrementa conforme pasa el tiempo y las decisiones se retrasan. Por lo cual, es fundamental equilibrar la correcci\'on de la clasificación y la rapidez en la detección de usuarios en riesgo. 

En los últimos años, se han propuesto diversos desafíos como CLEF eRisk~\cite{parapar2023} 
y MentalRiskES~\cite{marmol2023} que abordan problemas de DAR en distintos dominios vinculados a la salud mental. En particular, MentalRiskES 2023 fue la primera edición de DAR centrada en el idioma español, en la cual una de las tareas consistió en la detección de signos de depresión. El objetivo fue que los participantes propusieran soluciones para la DAR analizando los usuarios post a post.  Si un usuario es clasificado como positivo, se emite una alarma de riesgo y el análisis finaliza; en otro caso, si no se emiten alarmas durante todo el análisis, el usuario es clasificado como negativo. Las soluciones se evalúan utilizando métricas de clasificación estándar, como Accuracy, Precision, Recall y F1. Dado que el tiempo cumple un rol esencial en estos problemas, existen métricas específicas como \emph{early risk detection error} (ERDE)~\cite{losada2017}, que permite evaluar la precisión y la demora en las decisiones de los modelos. En particular, ERDE$\theta$ se distingue por penalizar severamente los casos verdaderos positivos (TPs) que ocurren después de haber leído $\theta$ posts. Sin embargo, la elección de este valor puede resultar en una métrica inadecuada, penalizando insuficientemente a los sistemas rápidos y de manera excesiva a los sistemas lentos~\cite{Sadeque2018}. De esta manera, surge la métrica F-latency~\cite{Sadeque2018}, que evalúa las soluciones utilizando la medida F1 y penaliza gradualmente las decisiones en función de la demora. 

Asimismo, el surgimiento de los Grandes Modelos de Lenguajes (LLMs) ha revolucionado el campo de la Inteligencia Artificial, demostrando una capacidad excepcional para resolver una amplia variedad de tareas en múltiples dominios. Su éxito se debe a la habilidad para aprender patrones complejos, entender expresiones en lenguaje natural y generar contenido de calidad y en múltiples formatos. El rendimiento general de los LLMs se evalúa usualmente mediante \emph{benchmarks} estandarizados. En particular, el modelo \emph{Gemini}~\cite{Gemini2023} ha demostrado un desempeño superior a otros modelos, como GPT-3.5 y LLAMA-2, en diversas pruebas como MMLU, GSM8K y BIG-Bench-Hard. Estos resultados destacan la capacidad de \emph{Gemini} para comprender y razonar en tareas lingüísticas complejas, convirtiéndolo en una herramienta valiosa para diversas aplicaciones. No obstante, para una comprensión más profunda de las ventajas y limitaciones de los LLMs, es importante evaluarlos en 
dominios específicos.

A pesar de su excelente rendimiento, los LLMs se consideran cajas negras debido a la falta de transparencia en sus procesos internos, lo que dificulta su interpretación y explicabilidad~\cite{EXP}. Asimismo, enfrentan desafíos importantes como las alucinaciones~\cite{HALLUCINATION1}, los sesgos~\cite{BIAS} y el alto costo computacional en entrenamiento e inferencia~\cite{COST}, que a menudo requiere el uso de APIs. De esta forma, el área de la interpretabilidad busca validar y comprender las decisiones de los LLMs, especialmente aquellas difícilmente explicables para los humanos~\cite{INTERPRETABILITY}. En esta área se realizan avances importantes para aumentar el grado de confianza de los LLMs, particularmente en aplicaciones críticas~\cite{RiskAnalysis}, donde los modelos deberían ser capaces de realizar predicciones correctas, como así también aportar información novedosa y relevante a profesionales~\cite{MED}. En~\cite{LLMyHealth} se presenta un estudio exhaustivo sobre los desafíos y oportunidades inherentes a la aplicación de los LLMs en el área de la psicología, enfocándose en aspectos como la precisión, eficacia y confiabilidad de los modelos. Aunque hay estudios que exploran el uso de los LLMs en diagnósticos médicos de pacientes~ \cite{DIAG1} 
y en la generación de respuestas automáticas para asistentes virtuales~ \cite{GENERATION2}, 
no se ha explorado el uso de los LLMs para resolver la DAR mediante razonamientos que respalden las soluciones obtenidas. 

Por otra parte, los LLMs generativos son modelos auto-regresivos entrenados para predecir una pieza de texto a partir de una secuencia de texto previa. 
La capacidad de generar explicaciones es una habilidad emergente desarrollada durante el entrenamiento~\cite{EMERGENTS}, que puede ser aprovechada para realizar predicciones precisas basadas en pasos de razonamiento explícitos~\cite{REASONING1}. Técnicas como cadena de pensamiento (\emph{chain-of-thought})~\cite{CoT1} y aprendizaje en contexto (\emph{in-context learning})~\cite{ICL1} 
pueden mejorar significativamente el desempeño de los modelos, generando respuestas fundamentadas y coherentes. En particular, \emph{in-context learning} es una técnica de pocos disparos (\emph{few-shot}) que permite a los modelos aprender patrones a partir de un número limitado de ejemplos en el \emph{prompt} de entrada. Aunque esta técnica puede mejorar la generación de razonamientos, requiere datos de entrenamiento bien fundamentados y no garantiza la precisión de los LLMs~\cite{STAR}. De esta forma, la evaluación de los LLMs en tareas de razonamiento se ha convertido en un área de gran interés para la comunidad científica~\cite{REASONING2}.

En este trabajo, proponemos un método interpretable para la DAR de depresión utilizando LLMs sobre textos en español. Nuestras principales contribuciones son: 1) definir un criterio de razonamiento para evaluar usuarios a través de un especialista (psicólogo), 2) aplicar la técnica \emph{in-context learning} al modelo Gemini para obtener predicciones fundamentadas, y 3) evaluar la solución obtenida de manera cuantitativa y cualitativa. Particularmente, comparamos los resultados obtenidos con métodos del estado del arte y evaluamos las respuestas con el especialista, quien observó contenidos relevantes y valiosos para la detección de usuarios con depresión. De esta manera, es posible obtener predicciones precisas sustentadas por razonamientos explicativos, permitiendo una comprensión más profunda del problema. Este estudio ofrece nuevas perspectivas para la DAR aprovechando el potencial de los LLMs.

\vspace{-2mm}

\section{Método}
Esta sección detalla los aspectos más importantes del método propuesto.

\vspace{-0.2cm}

\subsection{Razonamiento de muestras}
\begin{figure}[hbt]
\centering
\includegraphics[width=0.85\textwidth]{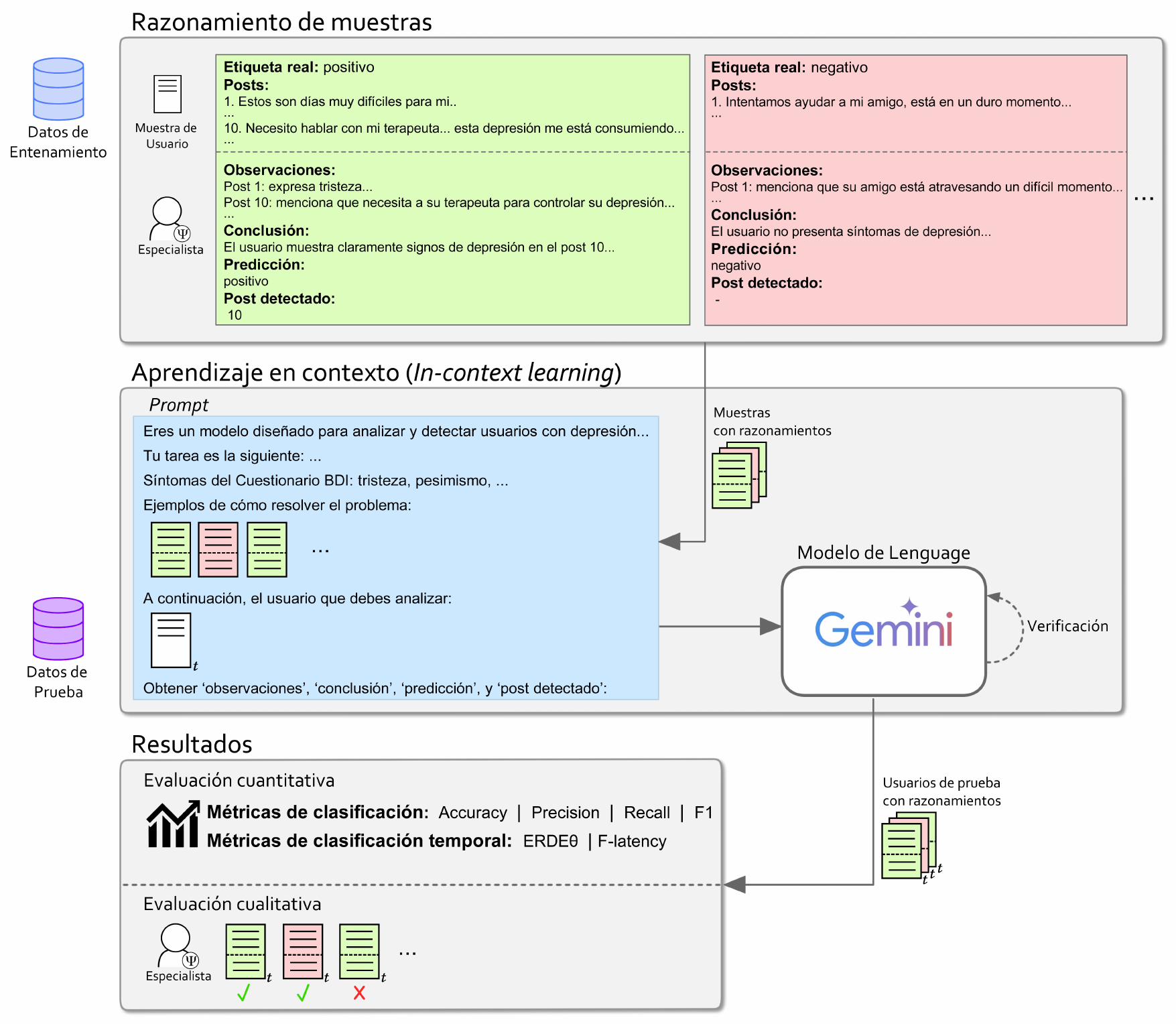} 
\caption{
En la etapa de razonamiento de muestras, un especialista analiza las muestras de entrenamiento según los síntomas del Cuestionario BDI, obteniendo observaciones, conclusión, predicción e identificación del post donde el usuario muestra signos de depresión. En la etapa \emph{in-context learning}, se construye un \emph{prompt} con las muestras más relevantes y el modelo Gemini evalúa los datos de prueba. Por último, los resultados se evalúan de forma cuantitativa y cualitativa.
}
\label{fig:scheme}
\end{figure}

En un problema típico de DAR, cada muestra de usuario consiste en una lista de posts y una etiqueta que indica si es positivo o negativo. Proponemos realizar una etapa de razonamiento de las muestras de entrenamiento siguiendo los criterios de un especialista en depresión.
El especialista analiza cada post utilizando los síntomas del Cuestionario BDI~\cite{MANUALBDI} para identificar posibles signos de depresión y elabora una lista de observaciones.
Luego, obtiene una conclusión y emite una predicción junto con el número de post en el que la información es suficiente para determinar que el usuario muestra claros signos de depresión. Por ejemplo, en la Fig.~\ref{fig:scheme} el especialista determinó que el primer usuario es positivo en el post 10 y el segundo es negativo, justificando ambas decisiones mediante las correspondientes observaciones y conclusiones. 

\vspace{-0.5cm} 
\vspace{-0.5cm} 


\subsection{Aprendizaje en contexto y evaluación de resultados}

Las muestras con razonamientos más relevantes son seleccionadas para construir el \emph{prompt}, mediante el siguiente esquema:

\medskip \noindent
\textbf{Rol.} Asignar inicialmente un rol al modelo ayuda a orientar el uso de sus capacidades y habilidades hacia la tarea a resolver. Por ejemplo: \emph{Eres un modelo diseñado para analizar y detectar usuarios con depresión}.
\vspace{-0.05cm} 

\medskip \noindent
\textbf{Definición de la tarea.} Especificar claramente el problema a resolver, detallando los pasos que el modelo debe seguir para solucionarlo. Por ejemplo:\\ 
\emph{
Tu tarea es la siguiente:\\
$i$. Lectura inicial de posts: lee la lista de posts del usuario para comprender el contexto general.\\
$ii$. Análisis y extracción de posts relevantes: analiza y extrae los posts donde el usuario presenta o expresa síntomas del Cuestionario BDI. \\
$iii$. Lista de observaciones: utiliza el Cuestionario BDI para construir observaciones, describiendo los posts extraídos y su vínculo con los síntomas del Cuestionario.\\
$iv$. Verificación: verifica y conserva los posts más relevantes.\\
$v$. Conclusión: elabora un breve resumen basado en las observaciones.\\
$vi$. Predicción: indica si el usuario es positivo o negativo para depresión.\\
$vii$. Post detectado: indica el número de post donde el usuario muestra claros signos de depresión.}

\vspace{-0.05cm} 

\medskip \noindent
\textbf{Ejemplos.} Se incluyen muestras con razonamientos para definir un contexto adecuado sobre cómo abordar la tarea. Los ejemplos más relevantes permiten al modelo ajustar su comportamiento durante la inferencia y generar predicciones precisas, rápidas y fundamentadas.
\vspace{-0.05cm} 

\medskip \noindent
\textbf{Consideraciones.} Se mencionan aspectos importantes como la lista de síntomas del Cuestionario BDI, medidas para evitar predicciones incorrectas, e incluso indicaciones para garantizar un formato de salida adecuado. 
\vspace{-0.05cm} 

\medskip \noindent
\textbf{Entrada.} Por último, se incluye la lista de posts del usuario a evaluar. 

\medskip \noindent
El modelo se invoca utilizando el \emph{prompt} para analizar los datos de prueba. Durante este proceso, llevamos a cabo una verificación para asegurar que todas las muestras sean evaluadas correctamente. De esta manera, para cada usuario, se obtienen las observaciones, una conclusión, la predicción y el número de post.

Los resultados generados por el LLM pueden ser evaluados cuantitativamente, usando métricas de clasificación estándar y temporales. Asimismo, es posible hacer una evaluación cualitativa, donde el especialista analice las decisiones y razonamientos realizados por el modelo.

\section{Resultados experimentales}
Los experimentos se realizaron sobre la DAR de depresión en idioma español, utilizando el corpus del desafío MentalRiskES 2023. 
Los organizadores obtuvieron los datos a partir de publicaciones de usuarios de la red social \emph{Telegram} y realizaron el etiquetado mediante anotadores reclutados en la plataforma \emph{Prolific}.
Los datos obtenidos no pueden ser divulgados públicamente, conforme a un acuerdo establecido entre los organizadores y los participantes (más detalles en~\cite{marmol2023}). El corpus incluye los datasets \emph{train}, \emph{trial} y \emph{test}, detallados en la Tabla~\ref{tab:depression_ds}. Utilizamos \emph{train} y \emph{trial} para la etapa de razonamiento, donde el especialista analizó 60 muestras y seleccionó las más relevantes para construir el \emph{prompt} de la etapa \emph{in-context learning}. Posteriormente, evaluamos nuestra propuesta con el conjunto \emph{test}.

\begin{table}[htb]
\vspace{-0.3cm} 
\caption{Detalles del corpus de depresión. Se muestra el número de usuarios (total, positivos y negativos) y posts por usuario (media, mínimo y máximo).}
\vspace{-1.5mm}
\label{tab:depression_ds}
\centering
\begin{tabular}{lcccccc}
\hline
\noalign{\vspace{0.3mm}}
\hspace{10mm} & \multicolumn{3}{c}{\textbf{\#Usuarios}} & \multicolumn{3}{c}{\textbf{\#Posts por usuario}} \\
\noalign{\vspace{0.3mm}}
 & \textbf{\hspace{2mm}Total\hspace{2mm}} & \textbf{\hspace{2mm}Pos\hspace{2mm}} & \textbf{\hspace{2mm}Neg\hspace{2mm}} & \textbf{\hspace{2mm}Media\hspace{2mm}} & \textbf{\hspace{2mm}Min\hspace{2mm}} & \textbf{\hspace{2mm}Max\hspace{2mm}} \\ 
\noalign{\vspace{0.3mm}}
\hline
\noalign{\vspace{0.3mm}}
Train & 175 & 94 & 81 & 35.7 & 11 & 100 \\
Trial & 10 & 6 & 4 & 62.4 & 11 & 100 \\
Test & 149 & 68 & 81 & 34.7 & 11 & 100 \\ 
\noalign{\vspace{0.3mm}}
\hline
\end{tabular}
\end{table}

Utilizamos el modelo \emph{Gemini-Pro}, configurando \emph{temperature}=0.2 y \emph{top\_p}=0.4 para asegurar un comportamiento mayormente determinístico. Los \emph{prompts} para evaluar las muestras del conjunto \emph{test} se mantuvieron dentro del límite de $32.000$ tokens del modelo. La implementación se llevó a cabo mediante la API de \emph{LangChain}\footnote{\url{https://python.langchain.com/v0.2/docs/introduction}} y se utilizaron las GPU proporcionadas por el entorno \emph{Google Colab}\footnote{\url{https://colab.research.google.com/}}.

\subsection{Resultados cuantitativos}
Utilizamos las respuestas del modelo para realizar una evaluación cuantitativa y medir su desempeño a través de diversas métricas. La Tabla \ref{tab:depression_results} compara nuestros resultados con las 5 mejores propuestas del MentalRiskES 2023, mostrando que nuestro modelo (InterpretableLLM\_Gemini) obtuvo el mejor rendimiento en todas las métricas. En términos de clasificación, superamos a UNSL\#1 en Accuracy, Precision, Recall y, especialmente en F1, logrando un 11\% más en esta métrica. 
En clasificación anticipada, nuestra propuesta superó a BaseLine-RoBERTaLarge\#1 en ERDE5, y a SINAI-SELA\#0 en ERDE30. Además, obtuvimos una mejora del 9\% en F-latency, superando a SINAI-SELA\#0 y BaseLine-Deberta\#0.
Estos resultados demuestran que el modelo adquirió la capacidad de tomar decisiones precisas y rápidas, resolviendo satisfactoriamente la tarea propuesta.

Por otra parte, de las 149 muestras del conjunto \emph{test}, se identificaron 63 TPs, 62 verdaderos negativos (TNs), 5 falsos negativos (FNs) y 19 falsos positivos (FPs). Se detectaron 2 muestras no procesadas por el modelo, posiblemente debido a restricciones éticas o de seguridad de la API utilizada, a las cuales se les asignó una predicción negativa por defecto. Cada muestra fue evaluada en 10 a 30 segundos, resultando en un tiempo total de aproximadamente 1 hora y 15 minutos para completar el análisis del conjunto \emph{test}.

\begin{table}[htb]
\caption{Resultados obtenidos considerando las métricas de clasificación (Accuracy, Precision, Recall y F1) y detección anticipada (ERDE5, ERDE30 y F-latency) para la tarea de Depresión. Se muestran los cinco mejores resultados según el ranking de los organizadores de MentalRiskES según ERDE30, así como los valores medios entre todos los resultados. Los valores en negrita y subrayados representan el primer y segundo mejor desempeño para cada métrica, respectivamente. 
Para ERDE$\theta$, valores cercanos a 0 representan un mejor desempeño, mientras que para las demás métricas, valores cercanos a 1 son preferibles.}
\vspace{-1.5mm}
\label{tab:depression_results}
\centering
\resizebox{\columnwidth}{!}{%
\begin{tabular}{lccccccc}
\noalign{\vspace{0.5mm}}
\hline
\noalign{\vspace{0.5mm}}
\textbf{Rank-Team\#Model} & \multicolumn{1}{l}{\textbf{Acc}} & \textbf{P} & \textbf{R} & \textbf{F1} & \textbf{ERDE5$\downarrow$} & \textbf{ERDE30$\downarrow$} & \textbf{F-latency} \\ 
\hline
\noalign{\vspace{1mm}}
1-SINAI-SELA\#0 \cite{gonzalez2023sinai} & 0.73 & 0.78 & 0.74 & 0.72 & 0.395 & {\ul 0.140} & {\ul 0.72} \\
\noalign{\vspace{0.3mm}}
2-UNSL\#1 \cite{thompson2023early} & {\ul 0.74} & {\ul 0.79} & {\ul 0.76} & {\ul 0.73} & 0.567 & 0.148 & 0.61 \\
\noalign{\vspace{0.3mm}}
3-BaseLine-Deberta\#0 \cite{marmol2023} & 0.66 & {\ul 0.79} & 0.69 & 0.64 & 0.303 & 0.153 & {\ul 0.72} \\
\noalign{\vspace{0.3mm}}
4-BaseLine-RoBERTaLarge\#1 & 0.70 & 0.76 & 0.72 & 0.69 & {\ul 0.290} & 0.159 & 0.70 \\
\noalign{\vspace{0.3mm}}
5-SINAI-SELA\#1 & 0.69 & 0.75 & 0.71 & 0.68 & 0.389 & 0.159 & 0.70 \\
\noalign{\vspace{0.3mm}}
\textit{MentalRiskES2023-mean} & 0.63 & 0.73 & 0.66 & 0.62 & 0.383 & 0.232 & 0.60 \\ 
\noalign{\vspace{0.3mm}}
\hline
\noalign{\vspace{0.5mm}}
InterpretableLLM\_Gemini & \textbf{0.84} & \textbf{0.85} & \textbf{0.85} & \textbf{0.84} & \textbf{0.262} & \textbf{0.105} & \textbf{0.81}\\
\noalign{\vspace{0.5mm}}
\hline
\vspace{-0.5cm} 
\end{tabular}
}
\end{table}

\subsection{Resultados cualitativos}
\begin{figure}[htb]
\centering
\vspace{-0.2cm} 
\includegraphics[width=0.9\textwidth]{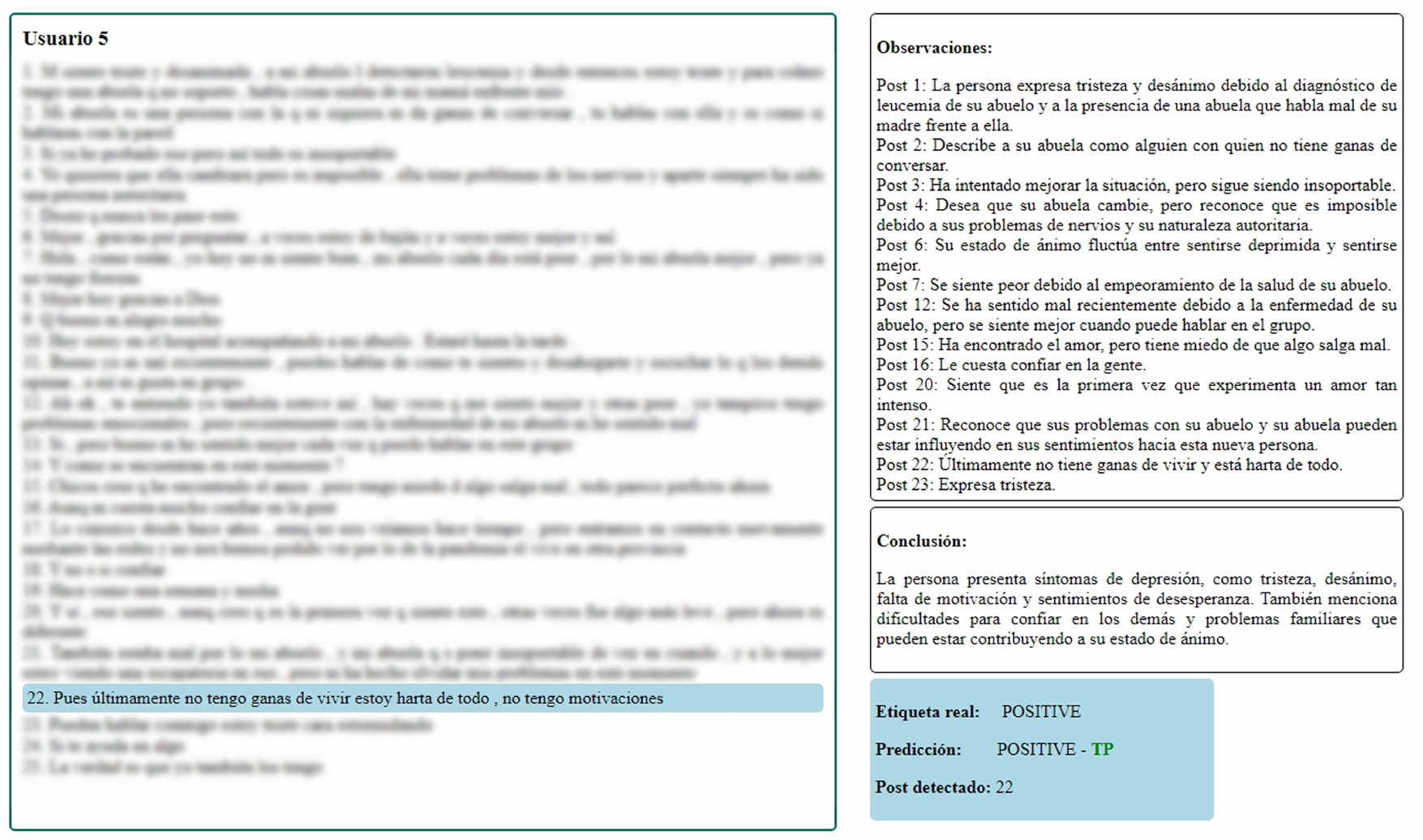} 
\caption{Herramienta para visualizar respuestas generadas por un LLM. En la columna izquierda se presenta la lista de posts del usuario (contenido difuminado para preservar el anonimato). La columna derecha muestra las observaciones y la conclusión del modelo, seguida por la etiqueta real del usuario (del conjunto \emph{test}) y la predicción, indicando si se trató de un caso TP, TN, FP o FN. Finalmente, se señala el número de post en el que se detectó al usuario (post detectado).
}
\vspace{-0.5cm} 
\label{fig:sample}
\end{figure}

El enfoque propuesto en este estudio permite que la solución sea interpretable. En esta sección, presentamos un análisis cualitativo del rendimiento del LLM en la tarea planteada.
Implementamos una herramienta para visualizar las respuestas generadas por el modelo (Fig.~\ref{fig:sample}) y solicitamos al especialista que evaluara los resultados en términos de relevancia y exactitud según los síntomas del Cuestionario BDI. 

\vspace{-0.25cm} 
\subsubsection{Predicciones correctas.}
El modelo es capaz de reconocer de forma acertada los principales síntomas de depresión. En los TPs, el modelo identifica rápidamente síntomas considerados críticos, como la ideación suicida y la desesperanza. 
Además, el modelo consideró referencias a emociones negativas como tristeza y llanto, destacando en algunos casos indicadores que refuerzan la gravedad de los síntomas, tal como la expresión \emph{sentimiento profundo de tristeza}. El modelo también detectó menciones explícitas sobre la patología, como afirmaciones de tener depresión, asistencia a terapia o descripciones de episodios personales. 
Por otra parte, se observó que el modelo utilizó una mayor cantidad de posts para tomar una decisión cuando un usuario planteaba mensajes contradictorios o hacía referencias terceros, antes de mostrar síntomas claros de depresión.
La detección también podía retrasarse ante mensajes indirectos sobre los síntomas; por ejemplo, un deseo de suicidio provocaba una alerta inmediata, mientras que si el usuario expresaba falta de ganas de vivir, la detección se demoraba esperando síntomas adicionales. En los TNs, el modelo fue capaz de detectar usuarios sin síntomas de depresión, generalmente indicando \emph{sin observaciones}. En algunos casos, describió brevemente los temas y comportamientos vistos en los posts y, en particular, generó observaciones cuando se mencionaban temas relacionados con la salud mental o sintomatología psicológica. En estos casos, el modelo justificó su diagnóstico proporcionando una conclusión explicativa razonable.
Además, resolvió adecuadamente situaciones potencialmente confusas, como cuando un usuario menciona que un tercero está atravesando un episodio depresivo o cuando describe experiencias pasadas de depresión sin mostrar síntomas actuales.

\vspace{-0.25cm} 
\subsubsection{Predicciones incorrectas.}
Es importante destacar que el modelo ha obtenido pocos FNs, dado el contexto crítico del problema. 
Aunque se asume que las etiquetas originales fueron correctamente asignadas, no siempre es posible identificar episodios depresivos con claridad en estos usuarios. El modelo demostró consistencia al detectar como negativos a usuarios con episodios pasados de depresión. 
Hay casos en que el usuario expresa sentimientos de tristeza, dificultad para dormir y malestar emocional, que el modelo pudo haber interpretado como insuficientes para el diagnóstico o relacionarlos con un carácter más somático. También mostró dificultades cuando el contenido es figurado. A pesar de esto, es meritorio que el modelo justifique las decisiones, aunque el diagnóstico sea presuntamente erróneo. 
Para los FPs, se observaron diferentes fallas en el razonamiento. En general, las justificaciones fueron incompletas o parcialmente correctas, como en casos donde se asignaron erróneamente síntomas no presentes en el usuario. También hubo errores al evaluar la intensidad de los síntomas para tomar una decisión. 
En algunos casos, el modelo confundió usuarios que han pasado episodios de depresión y actualmente brindan ayuda a otras personas, probablemente por las numerosas referencias a la depresión en sus posts.
Es importante destacar que muchos de estos casos son difíciles de analizar debido al contexto y a la cantidad limitada de información. Además, pueden existir diversas interpretaciones relacionadas a otros cuadros clínicos o características de personalidad, como irritabilidad y ansiedad.


\section{Conclusión y trabajo futuro}
En este trabajo, presentamos un método interpretable para la DAR de depresión utilizando LLMs sobre textos en español. 
Evaluamos muestras de usuarios con un especialista considerando síntomas del Cuestionario BDI y aplicamos la técnica \emph{in-context learning} al modelo Gemini para obtener respuestas fundamentadas. 
Nuestro enfoque obtuvo predicciones precisas y rápidas, con un desempeño notablemente superior a los métodos del estado del arte. Observamos que, definiendo adecuadamente los \emph{prompts} de entrada, es posible adaptar al modelo para generar razonamientos coherentes y valiosos en la detección de usuarios con depresión. Esto resalta la capacidad de los LLMs para resolver problemas complejos, proporcionando respuestas eficientes y justificadas. Sin embargo, la intervención de especialistas sigue siendo esencial para garantizar respuestas completamente correctas.

Los razonamientos obtenidos con los LLMs pueden significar recursos valiosos para resolver problemas de DAR. Es posible construir y mejorar conjuntos de datos donde las etiquetas estén adecuadamente justificadas. Esto permitiría incorporar un nuevo objetivo en los problemas de DAR: no sólo obtener respuestas correctas y rápidas, sino también asegurar una justificación adecuada para cada decisión. Además, se pueden desarrollar nuevas métricas de desempeño, como una versión mejorada de ERDE$\theta$ con umbrales individuales para cada usuario, evitando penalizaciones uniformes basadas en un único $\theta$. Por otra parte, sería interesante explorar otras técnicas para optimizar la adaptación de los LLMs en tareas de DAR, como el uso de grafos de conocimiento (\emph{Knowledge Graphs}) y \emph{Retrieval-Augmented Generation}. De esta manera, el estudio realizado en este trabajo representa un avance hacia nuevas perspectivas en la resolución de problemas de DAR.

\subsubsection*{Agradecidimientos.} 
Este trabajo forma parte de la tesis doctoral de Horacio Thompson y fue desarrollado en el Laboratorio de Investigación y Desarrollo en Inteligencia Computacional (LIDIC) [PROICO 03-0620] en la Universidad Nacional de San Luis (UNSL), Argentina.

\bibliographystyle{splncs03}

\end{document}